# Optimal Decomposition of Belief Networks

Wilson Xun Wen

Artificial Intelligence Systems,
Telecom Research Laboratories,
770 Blackburn Rd, Clayton,
Victoria 3168, Australia

**Abstract**

In this paper, optimum decomposition of belief networks is discussed. Some methods of decomposition are examined and a new method – the method of Minimum Total Number of States (MTNS) – is proposed. The problem of optimum belief network decomposition under our framework, as under all the other frameworks, is shown to be NP-hard. According to the computational complexity analysis, an algorithm of belief network decomposition is proposed in (Wen, 1989b) based on simulated annealing.

## 1 Introduction

The inherent disadvantage of probabilistic reasoning is an excessive requirement for computational resources (Cooper, 1987). There is an exponential explosion of the number of states as the number of variables in the probability space increases. One way to handle this problem is to decompose the underlying probabilistic space into small subspaces (Lemmer, 1983; Spiegelhalter, 1986), each of which corresponds to a hyperedge in an acyclic hypergraph, and calculate the probabilistic distributions of the small subspaces and propagate the changes among the subspaces through their intersections. Suppose the number of variables in the original space $S$ is $m$, and it has been decomposed into $n$ subspaces $S_i$, $1 \leq i \leq n$. If the $i$th subspace $S_i$ contains $m_i$ variables, then, in the case of binary variables, the number of states the probabilities of which are to be calculated is reduced from $2^m$ to $\sum_{i=1}^{n} 2^{m_i}$.

To see how huge a saving obtained by decomposition could be, let's have a look of a medium scale application with a probabilistic space of 24 binary variables (see Fig. 1 (a)). There are $2^{24} = 16,777,216$ states to be calculated without decomposition. Now suppose it is decomposed into $n = 11$ subspaces and all $m_i = 4$ (Fig. 1 (b)). Then there are only $11 \times 2^4 = 176$ states to be calculated after decomposition.

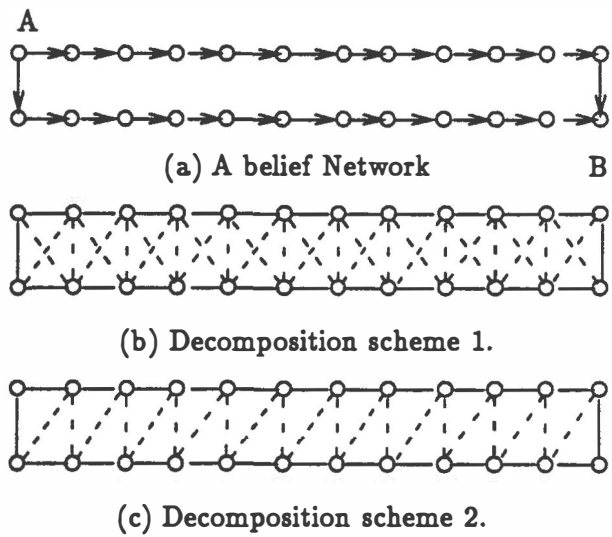

(a) A belief Network

(b) Decomposition scheme 1.

(c) Decomposition scheme 2.

Figure 1: Decomposition of a 24 variable space

Although the idea of decomposition of probabilistic space is very attractive, there remain technical issues to be discussed before the idea can be realized in practice. These issues are:

- How can we decompose the space into sub-

spaces so that the distributions of the subspaces, which are obtained by reasoning in the subspaces and propagation of the changes, exactly match with the marginal distributions of the result of reasoning in the original big space?

- How can we accomplish the biggest saving or how can we obtain the optimum decomposition?

- Because we also need to propagate the belief changes among the subspaces, there will be some delay caused by the propagation. How can we minimize such kind of delay?

- Is it possible to decompose a probabilistic space efficiently to satisfy the above requirements?

Several authors also discussed some of the above issues:

1. Lemmer (Lemmer, 1983) proposes an efficient algorithm to select "Component Marginal Distributions" (CMD) for an underlying distribution. This may be the earliest decomposition method for belief networks. Although this method only handles the case of belief tree networks, it provides a good methodology to reduce computational complexity in probabilistic reasoning.

2. Cheeseman (Cheeseman, 1983) proposes a reasoning method based on the Maximum Entropy Principle (MEP). To avoid an exponential explosion of the number of states as the number of variables increases, Cheeseman gives an efficient method to perform the relevant summations. The summations are divided into some subsummations. The complexity of the whole problem is reduced by this grouping significantly. It is easy to see that the basic idea behind this "grouping" is actually the same as decomposition.

3. Spiegelhalter (Spiegelhalter, 1986) uses an efficient "fill-in" algorithm (Tarjan and Yannakakis, 1984), to triangulate the underlying network. The decomposition consists of the maximal cliques of the triangulated network. This method can often obtain good results. For a comparison between Spiegelhalter's method and some reasoning method without decomposition such as Pearl's method (Pearl, 1986) in the case of multiconnected networks, see (Beinlich et al., 1989).

   Spiegelhalter's decomposition method pursues a result with minimum fill-in. Computing the minimum fill-in has been proven to be NP-complete (Yannakakis, 1981). That is to say, it is very hard to obtain the optimum result by existing methods. Furthermore, as pointed out by Kong (see the discussion in (Lauritzen and Spiegelhalter, 1988)), a result with a minimum fill-in is not necessarily the best result for the belief network decomposition.

4. Kong and Dempster (Kong, 1986; Dempster and Kong, 1988; Lauritzen and Spiegelhalter, 1988) propose searching for a fill-in which minimizes the maximal clique state size to triangulate the underlying networks. Although the maximal clique state size is vital to computational cost, the solution with the minimum maximal clique state size is still not necessarily the best. The problem to find a fill-in which minimizes the maximal clique state size has also been shown to be NP-complete (Arnborg et al., 1987).

It turns out that the trend in decomposition methods is towards the direction represented by Spiegelhalter's and Kong and Dempster's methods (Spiegelhalter, 1986; Kong, 1986; Dempster and Kong, 1988; Shenoy and Shafer, 1986; Shenoy and Shafer, 1988; Goldman and Rivest, 1988; Wen, 1990b). However, the criteria of the optimum decomposition for most of these methods are borrowed from some other areas (Tarjan and Yannakakis, 1984; Yannakakis, 1981; Arnborg et al., 1987). Others have discussed what is the most suitable criterion for the optimum decomposition of belief networks (see the discussion of (Lauritzen and Spiegelhalter, 1988)), but the essence of the problem – that the computa-tional amount of reasoning in belief network depends directly upon the total number of states in all subspaces of the decomposed network – has still not been caught directly. All analyses of the computational complexity of the problem of decomposition of belief networks are also borrowed from other areas (Yannakakis, 1981; Arnborg et al., 1987).

In this paper, we propose a decomposition method which directly minimizes the total number of states in the subspaces of the decomposed result. We show that the problem of optimum decomposition of belief networks under our framework, just like all the other frameworks, is NP-hard. Consequently, an algorithm of belief network decomposition is proposed in (Wen, 1989b) based on simulated annealing (Kirkpatrick et al., 1983) since it seems unlikely to develop a general efficient method in conventional way to solve an NP complete problem.

## 2  Belief Networks

In this section, following (Wen, 1989a), we introduce some concepts about belief networks based on the theory of Markov random fields (Kemeny et al., 1976) without any acyclic assumption.

Consider a probability space $X = \{x_i | i = 1, ... m\}$, with $M = a_1 \times ... \times a_m$ possible states $S = \{s_j | j = 1, ..., M\}$ and unknown probability distribution $p = \{P(s_j) | j = 1, ..., M\}$. Each variable $x_i$ in the space can take $a_i$ values. Suppose we have the following constraint set $CS$ on the distribution $p$ of $X$, which may be elicited from the domain expert or extracted from a sample database (Wen, 1990a):

1. Conditional constraints ($CCS$):

$$\mu_k = P(x_{k0} | X_{k1,...,kp_k}), \quad k = 1, ..., n \quad (1)$$

where $X_{k1,...,kp_k} = \{x_{k1}, ..., x_{kp_k}\} \subset X$.

2. Marginal constraints ($MCS$):

$$\nu_{k'} = P(X_{k'1,...,k'p'_{k'}}), \quad k' = 1, ... n' \quad (2)$$

For simplicity, we assume that these constraints can be added into the constraint set only when there are some corresponding conditional constraints (1) in $CS$ and

$$X_{k'1,...,k'p'_{k'}} \subseteq X_{k1,...,kp_k}.$$

although this assumption is not necessary for our final conclusion.

3. Universal constraint (UCS):

$$\sum_{x_0,...,x_{m-1}} P(x_0, ..., x_{m-1}) = 1.$$

This constraint is for the consistency. If we do not consider it as a constraint, then a normalization factor will be needed in the result.

According to the data dependencies and the conditional constraints in the constraint set, we may construct a directed graph, or *belief network* as follows

**Definition 2.1:** A *belief network* is a directed graph $G = (V, E)$, such that

1. The set of nodes of $G$ is $V = X$
2. The set of edges of $G$ is

$$E = \{< x_{kq}, x_{k0} >\}$$

such that $\exists \mu_k = P(x_{k0} | X_{k1,...,kp}) \in CS, x_{kq} \in X_{k1,...,kp}$.

Note that we do not make any independent or acyclic assumption here. $< x_i, x_j >$ is used to represent directed edge $x_i \rightarrow x_j$, and sometimes, we also use the corresponding undirected edge $(x_i, x_j)$ to represent both $< x_i, x_j >$ and $< x_j, x_i >$.

**Definition 2.2:** *Neighbor System of belief networks*

1. A *neighbor system* $\sigma$ in $G$ is a set of sets $\{\sigma x_i | x_i \in X, \sigma x_i \subseteq X\}$, such that
   (a) $x_i \notin \sigma x_i$,
   (b) $x_j \in \sigma x_i \iff \exists P(x_{k0} | X_{k1,...,kp}) \in CS, x_i, x_j \in X_{k0,...,kp}$

Obviously, we also have $x_i \in \sigma x_j \iff x_j \in \sigma x_i$.

2. The neighbors of a set $X' \subset X$ in G is the following set

$$\sigma X' = \{x_i \in X - X' | \exists x_j \in X, x_i \in \sigma x_j\}$$

3. The *neighborhood network* of a belief network $G = (V, E)$ is $G_\sigma = (V, E_\sigma)$, where

$$E_\sigma = \{(x_i, x_j) | x_i \in \sigma x_j\}.$$

Obviously we have $E \subseteq E_\sigma$ if we consider $E$ as a set of undirected edges.

4. A set $C \subseteq X$ is called a *clique* if $x_j \in \sigma x_i$ whenever $x_i, x_j \in C$ and $i \neq j$. A clique $MC$ is called *maximal clique* if there is no other clique $C$, such that $MC \subset C$. Let $CC$ and $MCC$ be the classes of all cliques and maximal cliques in $X$, respectively.

In (Wen, 1989a), some Markov Properties of general belief networks have been discovered.

**Theorem 2.1:** With the Maximum Entropy (ME) distribution (Cheeseman, 1983) $p = \{P(X)\}$ of $X$ subject to the constraint set $CS$, $p$ is a Markov random field (Kemeny et al., 1976) with respect to the neighbor system $\sigma$, and thus is also a neighbor Gibbs distribution (Kemeny et al., 1976) with respect to $\sigma$.

The ME distribution has quite a few desirable features for inductive reasoning and prediction (Rissanen, 1983) but it is difficult to obtain for big belief networks because there is an exponential explosion of the number of states as the number of variables increases. Furthermore, it is also very difficult to solve nonlinear programming problem to obtain the ME distribution. Therefore, it is always desirable

1. to look for close forms for some commonly encountered special cases of ME problems, such as marginal constraint problem and conditional constraint problem, etc.

2. to decompose big belief networks into small subnetworks and at the same time to maintain the consistency of the reasoning result among these subnetworks.

In the next section, we will briefly describe some basic techniques to decompose a belief network into an acyclic hypergraph, each of the hyperedges of which corresponds to a maximal clique in the extended neighborhood network. Issues like

1. how to obtain an ME distribution for the decomposed network,

2. how to guarantee and verify the consistency of the constraint set, and

3. why reasoning in the decomposed network is equivalent to reasoning in the original network

are discussed in (Wen, 1989a; Wen, 1990a).

## 3 Basic Techniques of Belief Network Decomposition

The concept of neighbor Gibbs field (see (Kemeny et al., 1976)) provides a valid factorization of the joint distribution for any Markov random field, so that it is possible for us to localize the computation of the joint ME distribution of the whole belief network within each of the maximal cliques of the neighborhood network (Definition 2.2). Actually, as we can see from definition 2.2 of neighbor system, each of the constraints on the underlying belief network is restricted within the corresponding clique of the neighborhood network. This suggests that the network should be decomposed into a hypergraph with the cliques of the neighborhood network as its hyperedges. To keep consistency among the distributions of the cliques, the reasoning results obtained in each clique need to be propagated to other cliques through their intersections. Consequently, it is desired to organize the decomposed result as an acyclic hypergraph (Beeri and Maier, 1981; Beeri et al., 1983) to guarantee the termination of the propagation and to avoid other possible anomalies during the propagation. This is not only necessary but also possible because any cyclic hypergraph can be converted into an acyclic one by expansion of its hyperedges, or equivalently by adding extra edges to its corresponding graph (Goldman and Rivest, 1988).

The basic techniques of belief network decomposition proposed in (Spiegelhalter, 1986; Lauritzen and Spiegelhalter, 1988; Kong, 1986; Dempster and Kong, 1988) can be described briefly as follows:

1. Construct a "moral graph", $G_\sigma = (V, E_\sigma)$ for belief network $G = (V, E)$, which is equivalent to neighborhood network of the belief network.

2. Find a filling-in (see (Tarjan and Yannakakis, 1984) and Definition 3.5) $F$ of $G_\sigma$, such that $D_\sigma = MCC_{G_f}$, the maximal clique set of $G_f = (V, F \cup E_\sigma)$, has

   - minimum $|F|$, for Spiegelhalter's method,
   - minimum $max\{MS_i\}$ for Kong and Dempster's method, where state size of $MC_i \in D_\sigma$:

   $$MS_i = \prod_{1 \leq j \leq \xi_i} |\eta_{ij}|,$$

   where $\xi_i$ is the number of variables in $MC_i$ and $\eta_{ij}$ is the number of values which can be taken by the $j$-th variable in $MC_i$,

$D_\sigma$ is the decomposition wanted and corresponds to an acyclic hypergraph $<V, D_\sigma>$.

## 3.1 Acyclic Hypergraphs and their desirabilities

Acyclic hypergraphs have been studied intensively in the area of relational databases (Beeri and Maier, 1981; Beeri et al., 1983):

**Definition 3.1:**

1. A *hypergraph* $H = (V, E_H)$ is a pair of a node set $V$ and hyperedge set $E_H = \{e_h | e_h \subseteq V\}$.

2. A hypergraph is said to be *reduced* if

   $$\forall e_h, e'_h \in E_H, e_h \not\subseteq e'_h.$$

   All hypergraphs discussed in this paper are reduced.

3. The *graph* $G(H)$ of a hypergraph $H$ is the graph $(V, E_h)$ with

   $$E_h = \{(u, v) | \exists e_h \in E_H, u, v \in e_h\}.$$

**Definition 3.2:**

1. A *path* in a graph $G = (V, E)$ is a sequence of distinct nodes $<v_0, ..., v_k>$ such that $(v_i, v_{i+1}) \in E$, for $i = 0, ..., k-1$.

2. A *cycle* is a path with $k \geq 1$ and $(v_k, v_0) \in E$.

3. A hypergraph is *conformal* if every clique of $G(H)$ is included in a hyperedge of $H$.

4. A graph is *chordal* if every cycle of length at least 4 has a chord, ie., an edge joining two nonconsecutive nodes on the cycle.

5. A hypergraph $H$ is *acyclic* if $H$ is conformal and $G(H)$ is chordal.

It has been noted that acyclic hypergraphs have many desirable properties. Because of these properties, it is possible to develop efficient (polynomial-time) algorithms for solving problems about acyclic hypergraphs that are NP-complete in the unrestricted case.

Suppose we have the following definitions in database theory (Ullman, 1982; Beeri and Maier, 1981; Beeri et al., 1983):

**Definition 3.3:**

1. A set of relations (may be infinite relations) $r_1, ..., r_n$ over sets of attributes $R_1, ..., R_n$ are *pairwise consistent* if for all $i, j \in \{1, ..., n\}$,

   $$\pi_{R_i \cap R_j}(r_i) = \pi_{R_i \cap R_j}(r_j).$$

   where $\pi$ is the projection operator.

2. $r_1, ..., r_n$ are *globally consistent* if

$$\exists r \text{ over } \cup_{j=1}^{n} R_j, \forall i \in \{1, ..., n\},$$

$$r_i = \pi_{R_i}(r).$$

**Definition 3.4:** We say $R_1, ..., R_n$ have the running intersection property if $R_i$'s can be ordered as $S_1, ..., S_n$ (called the *running intersection ordering*) such that

$$\forall i, \ 1 < i \leq n, \exists j_i < i, ((S_i \cap \bigcup_{j=1}^{i-1} S_j) \subseteq S_{j_i}) \quad (3)$$

Beeri et al (Beeri and Maier, 1981; Beeri et al., 1983) proved

**Theorem 3.1:** The following conditions about $R_1, ..., R_n$ are equivalent:

1. $< V, E_h >$ is an acyclic hypergraph, where

$$V = \bigcup_{i=1}^{n} R_i, \ E_h = \{R_1, ..., R_n\}.$$

2. Pairwise consistency is equivalent to global consistency for relation set $r_1, ..., r_n$ over $R_1, ..., R_n$.

3. $R_1, ..., R_n$ have the running intersection property.

4. Graham reduction:
   (a) delete attributes that appear in only one $R_i$.
   (b) delete $R_i$ if $\exists R_j \ (j \neq i, \ R_i \subseteq R_j)$.
   reduces $R_1, ..., R_n$ to nothing if applied repeatedly.

Statement 2 guarantees that the consistency, when calculating the distributions of every hyperedge separately, needs to be maintained only between pairs of the hyperedges. This is important because to verify a global consistency in the unrestricted case is actually NP-complete (Honeyman et al., 1980). Statement 3 gives us a proper updating order to obtain a consistent ME result for the whole decomposed network, and we have proven ((Wen, 1989a)) that the ME result obtained by local computation within each hyperedge and then propagation of the local result to the whole decomposed network iteratively matches exactly with the ME result globally calculated for the whole original network. Statement 4 gives a convenient way to check the acyclicity of a belief network. However, if a belief network is not acyclic, then we need to use the following ordering and fill-in methods to make it so.

### 3.2 Ordering and Fill-in Algorithms

In the rest of this section, we briefly describe some important ordering and fill-in algorithms.

**Definition 3.5:**

1. For a graph $G = (V, E_\sigma)$ with $|V| = m$, an *ordering* of $V$ is a bijection

$$\alpha : \{0, ..., m-1\} \leftrightarrow V.$$

2. The *fill-in* associated with this ordering is the set of edges $F(\alpha) = \{(u, v)\}$, where

$$(u, v) \notin E_\sigma \land \exists P_{u,v} \subseteq E_\sigma \cup F(\alpha),$$

$$\forall w \in P_{u,v}(\alpha(w) < \alpha(u) \land \alpha(w) < \alpha(v))\}.$$

3. The *elimination graph* of $G$ with ordering $\alpha$ is $G(\alpha) = (V, E_\sigma \cup F(\alpha))$.

4. If $F(\alpha) = \emptyset$, then $F$ is called an $\emptyset$ fill-in and $\alpha$ is an $\emptyset$ fill-in ordering.

It is proven in (Tarjan and Yannakakis, 1984) that

**Theorem 3.2:** A graph $G$ is chordal iff it has an $\emptyset$ fill-in ordering.

**Theorem 3.3:** An eliminated graph of $G$ with any ordering is chordal.

#### 3.2.1 Dynamic Programming

Given any criterion function (Tarjan and Yannakakis, 1984) $\Psi$ and a graph $G$ with $|V| = m$, using dynamic programming techniques in (Bertele and Brioschi, 1969b; Bertele and Brioschi, 1969a; Brioschi and Even, ) it is possible to find an ordering which minimize $\Psi$. However, both the computational complexity and the storage requirement of this algorithm increase as $2^m$.

### 3.2.2 Heuristic Algorithms

Suppose $\alpha$ is an ordering of $G = (V, E_\sigma)$ with $|V| = m$. The *i-th elimination graph* $G_i = (V_i, E_i)$ is the graph obtained by deleting the last $i$ nodes of $G$ in ordering $\alpha$ and all the edges incident from these nodes. Four commonly used heuristic algorithms are briefly described below. All of these algorithms have the basic structures similar to the following framework:

```
Algorithm 3.1:
Input: A graph G_0 = (V_0, E_0) (|V| = m).
Output: An ordering α of the nodes.
Method:
for (i = m - 1, G_0 = G; i ≥ 0; i--) {
    choose α(i) ∈ G_{m-i-1} satisfying
    one of the conditions given below;
    /* Break ties arbitrarily */
    V_{m-i} = V_{m-i-1} - {α(i)};
    E_{m-i} = E_{m-i-1} - {(α(i),v) ∈ E_{m-i-1}};
    G_{m-i} = (V_{m-i}, E_{m-i});
}
```

In principle, only the conditions for different algorithms are different and will be described below:

1. **Minimum Degree Algorithm:** Denote $d(\alpha(i))$ the degree of a node $\alpha(i)$ in $G_{i-1}$. We have the following condition for Minimum Degree Algorithm (Rose, 1972):

$$d(\alpha(i)) = min_{v \in V_{m-i-1}}\{degree(v)\}.$$

2. **Minimum Deficiency Algorithm:** The deficiency of a node $v$ in a graph $G = (V, E_\sigma)$ is the set of nodes in $V$

$$D(v) = \{(v_i, v_j) | v_i, v_j \in \sigma v \wedge (v_i, v_j) \notin E_\sigma\}$$

We have the following condition for Minimum Deficiency Algorithm (Rose, 1972):

$$D(\alpha(i)) = min_{v \in V_{m-i-1}}\{D(v)\}.$$

3. **Lexicographic Search Algorithm:** The next algorithm is called *lexicographic search algorithm* (Tarjan and Yannakakis, 1984). For each unnumbered node $v$, maintain a list $l(v)$ of the number of numbered nodes adjacent to $v$. For two unnumbered nodes $u$ and $v$, we say $l(u) \leq l(v)$ if $l(u)$ is lexicographically less than $l(v)$. For two list $l_1 = <a_0, ..., a_k>$ and $l_2 = <b_0, ..., b_l>$, $l_1 < l_2$ if for some $j$, $a_i = b_i$ for $i = 0, ..., j-1$ and $a_j < b_j$, or if $a_i = b_i$ for $i = 0, ..., k$ and $k < l$. We say $l_1 = l_2$ only when $k = l$ and $a_i = b_i$ for $i = 0, ..., k$.

The condition for this algorithm (Lexicographic Search Algorithm) is as follows:

$$l(\alpha(i)) = max_{v \in V_{m-i-1}}\{l(v)\}.$$

4. **Maximum Cardinality Search Algorithm:** Defining the cardinality of an unnumbered node $u$ as

$$c(u) = max\{\alpha(v) | v \in \sigma u \wedge v \notin V_{m-i-1}\}$$

where $v \in \sigma u$ is in the original $G$ and it has been eliminated for the current eliminated graph, we have the following condition for the maximum cardinality search algorithm (Tarjan and Yannakakis, 1984):

$$c(\alpha(i)) = max_{v \in V_{m-i-1}}\{c(v)\}.$$

## 4 MTNS Decomposition

Using Spiegelhalter or Kong's method to triangulate the belief network in Fig. 1 (a), we obtain the decomposition shown in Fig. 1 (c). Comparing the two decomposition schemes in Fig. 1 (b) and (c), it is easy to see that

- Both decompositions are acyclic hypergraphs, and the number of filling-in edges in Fig. 1 (b) is much greater than that in Fig. 1 (c). Thus, according to Spiegelhalter's method, scheme 2 is better than scheme 1.

- The maximal clique state sizes for the case of binary variables are 16 (Fig. 1 (b)) and 8 (See Fig. 1 (c)), respectively. Thus, according to Kong's method, scheme 2, again, is better than scheme 1.

- However, from the point of view of the computational amounts of the two schemes, the above conclusions are not necessarily true: the total numbers of states the probabilities of which are to be calculated for the two schemes are the same – 176!

- More interestingly, the maximum propagation delay for scheme 2 (22 levels) is much greater than that for scheme 1 (11 levels)! That is to say, with the parallel reasoning model proposed in (Wen, 1988; Wen, 1989c), in an environment of a parallel computer system with at least 16 processors, scheme 1 is as nearly twice fast as scheme 2.

The above observation caused us to propose (Wen, 1990b) that a nearly optimum solution of decomposition should minimize the following amount (in the case of binary variables)

$$\sum_{i=1}^{n} 2^{m_i}, \quad (4)$$

where $n$ is the number of cliques (or hyperedges) in the network obtained by filling-in, $m_i$ is the number of variables in the i-th clique. This is called the method of the Minimum Total Number of States (MTNS), because it is easy to see that (4) is actually the total number of states to be calculated after decomposition. Keeping this amount minimum, our goal is to minimize the number of levels of propagation delay.

For the case of non-binary variables, the amount that should be minimized is

$$\sum_{i=1}^{n} \prod_{j=1}^{\xi_i} \eta_{ij}, \quad (5)$$

where $\eta_{ij}$ is the number of values can be taken by the $j$-th variable in the $i$-th clique and there are $\xi_i$ cliques in total. In the rest of this paper, we only discuss the case of binary variables but our result can be generalized to the case of non-binary variables straightforwardly.

## 5 Optimum Decomposition of Belief Network is NP-hard

Optimum decomposition of belief network can be formulated into the following problem:

**Instance:** Graph $G = (V, E)$, non-negative integer $k$.

**Question:** Is there a fill-in, $F = \{(u,v)|u,v \in V \wedge (u,v) \notin E\}$, with the total number of states less than or equal to $k$? That is, the addition of this fill-in makes the graph $G_f = (V, E \cup F)$ chordal and the total number of states in the set of maximal cliques in $G_f$:

$$D_\sigma = \{MC_i | MC_i \in MCC_{G_f}\}.$$

is

$$\sum_{i=1}^{n_\sigma} 2^{|MC_i|} \leq k, \quad where \quad n_\sigma = |D_\sigma|.$$

We are going to prove this problem is NP-hard.

**Definition 5.1:** ((Yannakakis, 1981)) A bipartite graph $G' = (P \cup Q, E')$ is a *chain graph* if the neighbors of the nodes in $P$ form a *chain*, ie. there is a bijection $\pi : \{1, 2, ..., |P|\} \leftrightarrow P$, such that $\sigma(\pi(i)) \supseteq \sigma(\pi(j))$ iff $i < j$, where $\sigma(v)$ is the neighbor set of $v$ in $G'$. The neighbors of the nodes in $Q$ also obviously form a chain.

Suppose graph $C(G') = (P \cup Q, E'')$ is constructed from $G'$ by making $P$ and $Q$ cliques, ie.

$$E'' = E' \cup \{(u,v)|u,v \in P\} \cup \{(u,v)|u,v \in Q\}.$$

**Lemma 5.1:** ((Yannakakis, 1981)) Let $G'$ be a bipartite graph. $C(G')$ is chordal iff $G'$ is a chain graph.

**Theorem 5.2:** Optimum decomposition of belief networks is NP-hard.

**Proof:** The reduction is from the *Elimination Degree Sequence (EDS) problem* (see p201, (Garey and Johnson, 1979)):

**Instance:** Graph $G = (V, E)$, sequence $< d_1, d_2, ..., d_m >$ of non-negative integers not exceeding $m - 1$, where $m = |V|$.

**Question:** Is there an one-to-one function $\tau : V \to \{1, 2, ..., m\}$ such that, for $1 \le i \le m$, if $\tau(v) = i$ then there are exactly $d_i$ nodes $u$ such that $\tau(u) > i$ and $(u, v) \in E$?

EDS problem has been shown to be NP-complete (Garey and Johnson, 1979).

Let $(G = (V, E); < d_1, d_2, ..., d_m >)$ be an instance of the EDS problem. Without loss of generality, suppose $G$ is connected. Construct a bipartite graph $G' = (P \cup Q, E')$ as follows:

- $P = V$,
- $Q$ has one node $e_n$ for each edge $e$ of $G$, and
- if $e = (u, v) \in E$ then node $e_n$ is adjacent to $u$ and $v$ in $G'$.

An example similar to that in (Yannakakis, 1981; Arnborg et al., 1987) is given in Fig. 2:

Set

$$k_i = \begin{cases} 0, & d_i = 0 \\ 2^{m-i+1+\sum_{j=1}^{i} d_j}, & d_i \ne 0 \end{cases} \quad (6)$$

and

$$k = \sum_{i=1}^{m} k_i, \quad (7)$$

where k does not actually depend on any ordering of the nodes in $V$. Suppose $\tau$ is an ordering of $V$ which satisfies the EDS problem. Choosing the reversed ordering $\pi$ of $\tau$, we construct a fill-in $F(\pi)$ of $C(G')$ which is uniquely specified by $\pi$ in the following way:

**P1:** For each node $x$ in $Q$, let $\delta(x) = max\{i|(x, \pi(i)) \in E'\}$.

**P2:** Then $F(\pi) = \{(x, \pi(j))|x \in Q, j < \delta(x), (x, \pi(j)) \notin E'\}$.

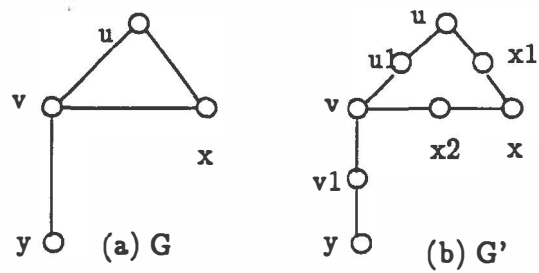

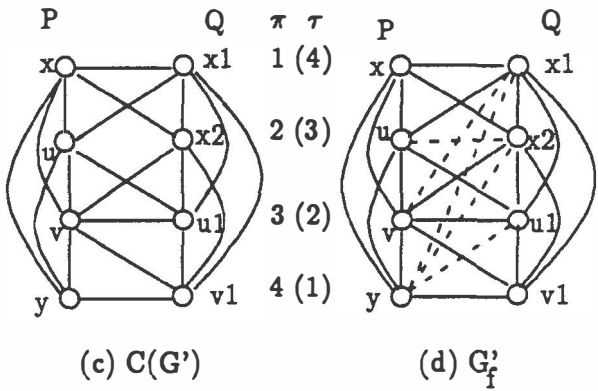

(c) C(G')  (d) $G'_f$

Figure 2: An example of ordering and fill-in

We claim that $F(\pi)$ satisfies the requirement of the corresponding optimum belief network decomposition for $C(G')$.

It is not very difficult to see that the addition of $F(\pi)$ to $C(G')$ makes the result graph $G'_f$ a chain graph and thus a chordal graph. To see this, suppose $i < j$, $\exists y \in Q$, $(y, \pi(j)) \in E' \cup F$, and $\sigma$ is the neighbor system in $G'_f$, then we have

- $j \le \delta(y)$ because of **P1**.
- Thus $i < \delta(y)$ and $(y, \pi(i)) \in E' \cup F$ because of **P2**.
- Therefore, $\sigma(\pi(i)) \supseteq \sigma(\pi(j))$.

Furthermore, the set of maximal cliques $D_\sigma$ of $G'_f$ which contains $m - z$ cliques, where $z$ is the number of $d_i$'s which equal to 0. The clique in $D_\sigma$ corresponding to the $i$th node in P for ordering $\tau$ contains

$$k_i = m - i + 1 + \sum_{j=1}^{i} d_j = m + 1 + \sum_{j=1}^{i}(d_i - 1)$$

nodes, among which $m - i + 1$ nodes are in $P$ and $\sum_{j=1}^{i} d_j$ nodes are in $Q$. However, if $d_i = 0$, then the corresponding clique is not maximal and thus should be excluded from the consideration. Thus we have the total number of states for $D_\sigma$ is exactly $k$ (see (6) and (7)).

To see this, consider that

- For $i = 1$, $\tau^{-1}(1)$ has all $m-1$ nodes in $P$ as its neighbors because we have made $P$ a clique. The $d_1$ nodes in $Q$, which correspond to the edges in $G$ with $\tau^{-1}(1)$ as one of its terminate nodes, are also the neighbors of $\tau^{-1}(1)$. $\tau^{-1}(1)$ has only these neighbors in $G'_f$, and the size of the neighborhood set of $\tau^{-1}(1)$ in $G'_f$:

$$|\sigma(\tau^{-1}(1))| = m - 1 + d_1.$$

Furthermore, all of $m+d_1$ nodes in $\sigma(\tau^{-1}(1)) \cup \{\tau^{-1}(1)\}$ are connected to each other because

- $P$ and $Q$ are cliques in $G'_f$,
- Each node $z$ in $\sigma(\tau^{-1}(1))$ is connected to $\tau^{-1}(1) = \pi(n)$. But we have $\forall y \in P$, $(z,y) \notin E' \to (z,y) \in F(\pi)$ because of P2. Therefore $(z,y) \in E' \cup F$.

Thus $\sigma(\tau^{-1}(1)) \cup \{\tau^{-1}(1)\}$ is a clique of $G'_f$. $\sigma(\tau^{-1}(1)) \cup \{\tau^{-1}(1)\}$ is also a maximal clique of $G'_f$ because

- $\tau^{-1}(1)$ is not a terminate node for any filling-in edges.
- All nodes in $P$ and $\sigma(\tau^{-1}(1))$ have been included in $\sigma(\tau^{-1}(1)) \cup \{\tau^{-1}(1)\}$.

- Suppose for $i$, $1 \leq i < m$, $S_i = \sigma(\tau^{-1}(i)) \cup \{\tau^{-1}(i)\}$ is a maximal clique in $G'_f$ with $|S_i| = m - i + 1 + \sum_{j=1}^{i} d_j$, and has $m - i + 1$ nodes in $P$ and $\sum_{j=1}^{i} d_j$ nodes in $Q$.

- For $i+1$, we have

  - $\tau^{-1}(i+1)$ has $m-i-1$ nodes $\tau(i')$ ($i' > i+1$) in $P$ as its neighbors.
  - $\tau^{-1}(i + 1)$ has all $\sum_{j=1}^{i} d_i$ nodes in $\sigma(\tau^{-1}(i)) \cap Q$ as its neighbors, because $G'_f$ is a chain graph and $\pi$ is the reversed ordering of $\tau$.
  - $\tau^{-1}(i+1)$ has $d_{i+1}$ extra neighbors in $Q$ corresponding to the edges

    $$(\tau^{-1}(i+1), \tau^{-1}(i')) \in E, \ i' > i + 1$$

    if $d_{i+1} \neq 0$. (Denote the set of these nodes by $S_{d_{i+1}}$.)

  All of these $m - i + \sum_{j=1}^{i+1} d_j$ nodes in $S_{i+1} = \sigma(\tau^{-1}(i+1)) \cup \{\tau^{-1}(i+1)\}$ are connected to each other in $G'_f$ because

  - $P$ and $Q$ are cliques in $G'_f$.

  - All nodes in $S_{i+1} \cap Q$ are connected to $\tau^{-1}(i + 1)$, thus also connected to all nodes $\tau^{-1}(i')$, $i' > i + 1$ because of P2.

  This clique is obviously maximal if $d_{i+1} \neq 0$, because

  - The nodes in $S_{d_{i+1}}$ are not connected to node $\tau^{-1}(i''')$, $i''' \leq i$.
  - If $\exists d_{i'} \neq 0$, $i' > i + 1$, then $\tau^{-1}(i+1)$ is not connected to all nodes in $S_{d_{i'}}$.

$G'_f$ only has the above maximal cliques. □

The numbers of filling-in edges, maximal clique state sizes, $d_i$'s, the numbers of total states and the delay levels for three ordering schemes in Fig. 2 and 3 are given in Table 1. It turns out that although the computational amount of a scheme has some relationship with both the number of filling-in edges and the maximal clique state size, however, this amount mainly depends on the total number of states the probabilities of which are to be calculated. Based on this observation, we adopt the schemes with the minimum total number of states (4) or (5) as the optimum decomposition schemes. In an environment of parallel computer systems, we also consider to adopt the scheme with the minimum number of delay levels for belief propagation. For example, suppose there are at least 16 processors in a parallel system. We may prefer the scheme in Fig. 1 (b) rather than 1 (c). We obviously need a tradeoff between some schemes like those in Fig. 3 (a) and (b).

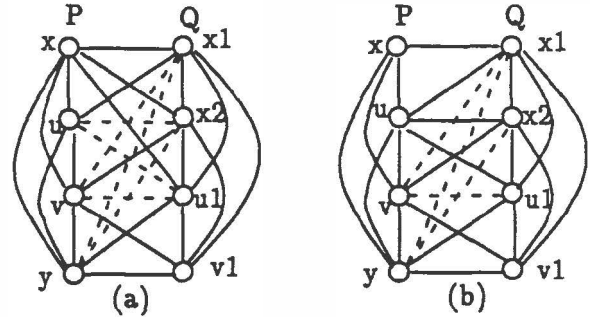

Figure 3: More orderings/fill-in for Fig. 2

| Scheme | 2 (d) | 3 (a) | 3 (b) |
|---|---|---|---|
| Fill-in | 5 | 6 | 4 |
| $M_{clique}$ | 6 | 7 | 6 |
| $d_i$'s | 2, 1, 1, 0 | 3, 0, 1, 0 | 1, 2, 1, 0 |
| $N_{states}$ | $3 \cdot 2^6$ = 192 | $2^7 + 2^6$ = 192 | $2^5 + 2 \cdot 2^6$ = 160 |
| Delay | 3 | 2 | 3 |

Table 1: Comparison for example in Fig. 2 & 3

## 6 Conclusions

To reduce the computational complexity in probabilistic reasoning, a decomposition of the underlying belief network is always desired. Several methods of belief network decomposition are examined and a new criterion – the MTNS criterion – of optimum results is proposed according to the computational amount required by the decomposed belief network during probabilistic reasoning. It is shown that the problem of optimum belief network decomposition under our framework, like some other frameworks, is NP-hard. To obtain an optimum or suboptimum decomposition of belief networks a new algorithm is also proposed in (Wen, 1989b) based on simulated annealing.